\documentclass[10pt,twocolumn,letterpaper]{article}

\usepackage[pagenumbers]{cvpr} % To force page numbers, e.g. for an arXiv version

\usepackage{graphicx}
\usepackage{amsmath}
\usepackage{amssymb}
\usepackage{booktabs}
\usepackage{xcolor,color,colortbl}

\usepackage[pagebackref,breaklinks,colorlinks]{hyperref}

\usepackage[capitalize]{cleveref}
\crefname{section}{Sec.}{Secs.}
\Crefname{section}{Section}{Sections}
\Crefname{table}{Table}{Tables}
\crefname{table}{Tab.}{Tabs.}
\newtheorem{myDef}{\textit{Definition}}

\def\cvprPaperID{*****} % *** Enter the CVPR Paper ID here
\def\confName{CVPR}
\def\confYear{2023}

\def\cvprPaperID{***} % *** Enter the CVPR Paper ID here
\def\confName{CVPR}
\def\confYear{2023}

\begin{document}

%%%%%%%%% TITLE - PLEASE UPDATE
\title{Masked Reconstruction Contrastive Learning with \\ Information Bottleneck Principle}

\author{Ziwen Liu\\
University of Chinese Academy of Sciences\\
{\tt\small liuziwen18@mails.ucas.ac.cn}
\and
Bonan Li\\
University of Chinese Academy of Sciences\\
{\tt\small libonan@ucas.ac.cn}
\and
Congying Han \thanks{Corresponding author}\\
University of Chinese Academy of Sciences\\
{\tt\small hancy@ucas.ac.cn}
\and
Tiande Guo\\
University of Chinese Academy of Sciences\\
{\tt\small tdguo@ucas.ac.cn}
\and 
Xuecheng Nie\\
MT Lab, Meitu Inc., China\\
{\tt\small nxc@meitu.com}
}

\maketitle

%%%%%%%%% ABSTRACT
\begin{abstract}
Contrastive learning (CL) has shown great power in self-supervised learning due to its ability to capture insight correlations among large-scale data. Current CL models are biased to learn only the ability to discriminate positive and negative pairs due to the discriminative task setting. However, this bias would lead to ignoring its sufficiency for other downstream tasks, which we call the discriminative information overfitting problem. In this paper, we propose to tackle the above problems from the aspect of the Information Bottleneck (IB) principle, further pushing forward the frontier of CL. Specifically, we present a new perspective that CL is an instantiation of the IB principle, including information compression and expression. We theoretically analyze the optimal information situation and demonstrate that minimum sufficient augmentation and information-generalized representation are the optimal requirements for achieving maximum compression and generalizability to downstream tasks. Therefore, we propose the Masked Reconstruction Contrastive Learning~(MRCL) model to improve CL models. For implementation in practice, MRCL utilizes the masking operation for stronger augmentation, further eliminating redundant and noisy information. In order to alleviate the discriminative information overfitting problem effectively, we employ the reconstruction task to regularize the discriminative task. We conduct comprehensive experiments and show the superiority of the proposed model on multiple tasks, including image classification, semantic segmentation and objective detection.
\end{abstract}

%%%%%%%%% BODY TEXT
\section{Introduction}
\label{sec:intro}

% 简单介绍CL，并强调它的重要性
Contrastive Learning (CL) is a fundamental and important problem in the self-supervised learning field. It aims to maximize the similarity of positive pairs and dissimilarity of negative ones without requirements of explicit data labeling. Due to the strong capability of scaling up training set and extracting representative features, it has shown promising results in a variety of tasks, such as image classification~\cite{yang2022unified,he2020momentum}, semantic segmentation~\cite{wang2021exploring,zhao2021contrastive} and object detection~\cite{xie2021detco,yang2021instance}. 
\begin{figure}[t!]
    \centering
    \includegraphics[width=\columnwidth]{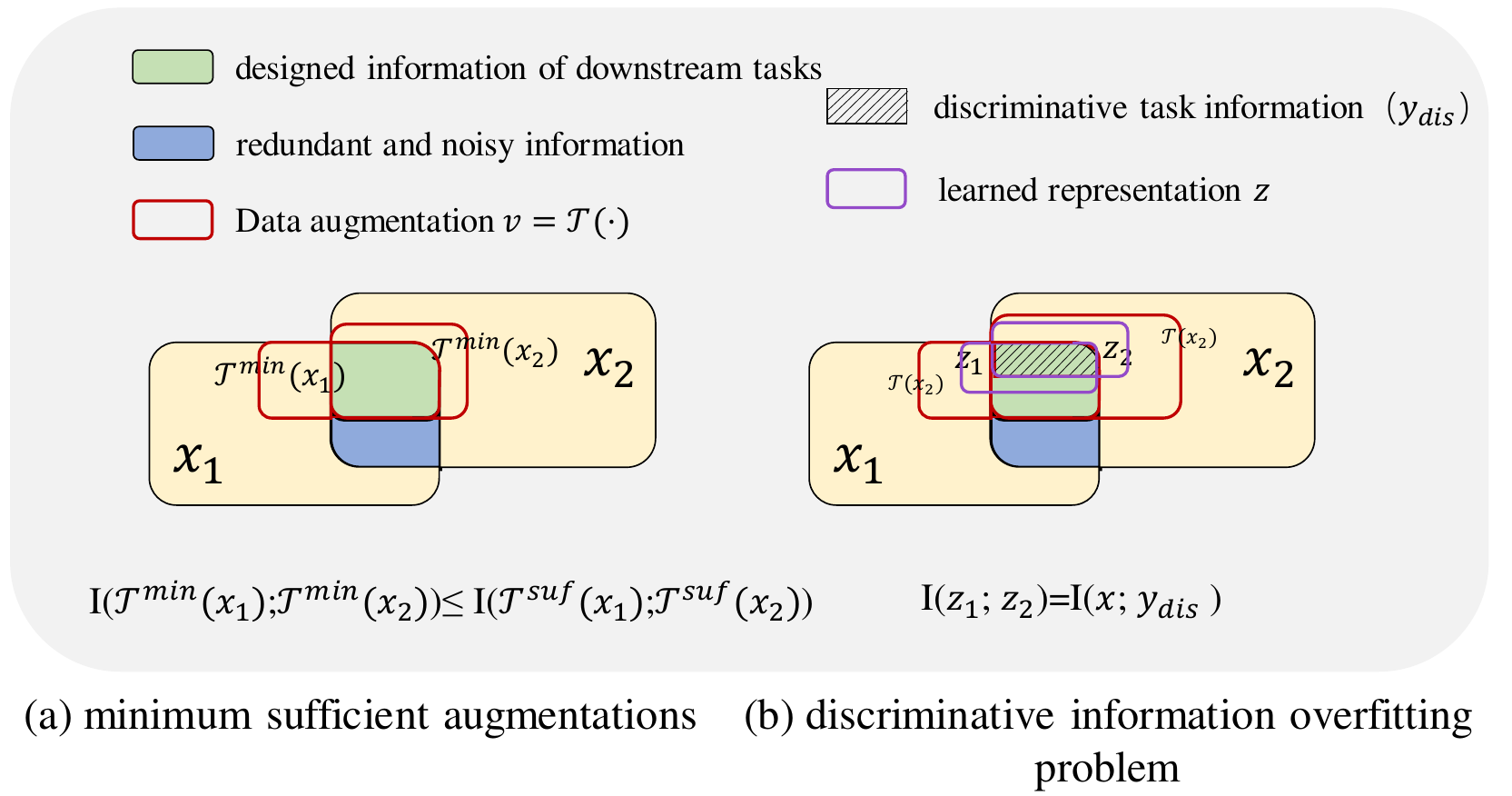}
    \caption{Motivation of our MRCL model. (a) We propose to exhibit an minimum sufficient augmentation that only compresses redundancy and remains essential information; (b) Only discriminative pretext task leads to the discriminative information overfitting problem, which degrades the performance on the various downstream tasks.}
    \label{fig:motivation}
\end{figure}

% 说明已有工作的实现并引出它们的问题，即我们在论文中要解决的问题。说明已有工作是怎么解决CL这个问题的，它们的好处是什么，同时，现有工作的问题又在哪里，这些问题的重要性和迫切性
% Existing models for CL solve this learning task via XXX, 
% Their advantages are XXX. due to the reasons...However, they suffer from the problems of XXX. These problems are caused by XXX, limiting the performance of CL in practice.

Existing works~\cite{chen2020simple,bachman2019learning,tian2020makes} on CL explore data augmentation to enhance generalizability and verify that a strong augmentation method can benefit the performance. However, due to the low information density of image data~\cite{he2022masked,xie2022simmim}, there is still much noisy and redundant information after their proposed augmentation, which would affect the extraction of desired information and thus limit the performance of CL in practice. Therefore, we need stronger information compression methods to yield~\textit{minimum sufficient augmentation} that can remove all redundant information.

In addition, for the pretext task setting, current CL models choose the discriminative task, which aims to measure and separate the distances of positive and negative samples to distinguish different samples. To keep the simplicity and efficacy, they focus on the discrimination of samples at the abstract semantic level of the feature space rather than complicated details.
However, the isolated discriminative task leads the model to fall into the bias that only extracts the information to help discriminate samples, i.e.,~\textit{discriminative information overfitting} problem, thus degrading the performance on different downstream tasks. 

% 说明我们的动机。重点指出为什么选择我们在论文中提出的方法来解决之前的问题。

Motivated by this, we propose to tackle the above problems from the perspective of the Information Bottleneck~(IB) principle~\cite{tishby2015deep,tishby99information,shwartz2017opening}, further pushing forward the frontier of CL. In particular, the IB principle aims to analyze information changes in the learning process of deep neural networks, relying on the idea of the simultaneous existence of information compression and information expression.
It has been widely applied to interpretability and engineering deep learning~\cite{alemi2017vib,voloshynovskiy2020variational,wu2020graph,wan2021multi,gao2021information}. Given this inspiration, we utilize IB principle to make theoretical analysis on the information variation and demonstrate that CL is actually an instantiation of the IB principle. We apply this principle at each processing stage for extracting the optimal information, thus leading to improved performance of CL.

% 基于以上动机，我们是怎么设计模型的，这里需要从较为high-leve的角度解释一下。
Specifically, this paper presents a novel Masked Reconstruction Contrastive Learning (MRCL) model. To compress the redundant information, existing works exploit data augmentation transformer to generate views and then further derive compact feature information shared by all views with multi-view architectures. However, they suffer from excessive redundancy problems when meeting image data with low information density.
Differently, our MRCL model conducts redundant information removal at the pixel level to attain the minimum sufficient augmentation. As shown in Fig.\ref{fig:motivation}(a), minimum sufficient augmentation can generate multiple views that share exactly the required downstream task information content. In addition, information expression of existing CL models is determined by the pretext task, \emph{i.e.,} discrimination task. This setting makes the model biased to extract the discriminative information, resulting in the overfitting problem and poor generalizability to different downstream tasks, as shown in Fig.\ref{fig:motivation}(b). To solve this problem, our MRCL model rectifies the information content expressed by representation with more generalized and critical information. In this way, MRCL is enhanced in generalizability. By combining the designs in terms of both redundant information compression and better information expression, our MRCL model can achieve a better information state. 

% 基于以上high-level的解释，在这里说明我们方法的具体实现，比如你提到的augmentation，mask方法、reconstruction term、loss、网络等。
For implementation, MRCL combines the mask operation with the current data augmentation mechanism in CL to enhance pixel level compression. Especially this masking operation has been verified to be effective in Masked Image Modeling~(MIM) and Masked Language Modeling~(MLM)~\cite{he2022masked,xie2022simmim,devlin2018bert,brown2020language}. The proposed new module can produce stronger augmentation results, thus improving the results\cite{chen2020simple,bachman2019learning}. In addition, MRCL introduces a reconstruction task as a regularization to make corrections on the information expression, avoiding the information bias and alleviating the overfitting problem. The backbone design of the encoder is also a point of concern. Following prior works, MRCL adopts a siamese structure encoder~\cite{bromley1993signature,chen2020simple,grill2020bootstrap,he2020momentum,chen2020improved,chen2021empirical,zbontar2021barlow,chen2021exploring} to learn the representations. In addition, MRCL applies Vision Transformer (ViT)~\cite{dosovitskiy2020image,liu2021swin} as the backbone, in light of its excellent representation learning capability for image data.

We conduct comprehensive experiments on multiple tasks to verify the efficacy of the proposed model, including image classification, semantic segmentation, and object detection. Results show that our MRCL model achieves superior performance over the prior state-of-the-art on multiple benchmarks (ImageNet1K~\cite{deng2009imagenet}, ADE20K~\cite{zhou2019semantic} and MSCOCO~\cite{lin2014microsoft}). In summary, our contributions are fourfold: 1) We present a new perspective on CL using the information bottleneck principle and theoretically analyze the information compression and expression therein. To our best knowledge, we are the first to point out problems of inadequate compression and discriminative information overfitting in CL. Given this, we define the minimum sufficient augmentation and information-generalized representation.
2) We present a novel combination of masking operation and basic data augmentation for deriving strong augmentation results, improving the performance of contrastive learning. 
3) We present an effective reconstruction task to regularize the discriminative task, leading to improved generalizability.
4) With the proposed model, called MRCL, we set new state-of-the-art for multiple tasks on multiple benchmarks. We also provide extensive analytical experiments to understand further our theoretical analysis and verify the rationality and effectiveness of our proposed components.

%-------------------------------------------------------------------------

\section{Related Work}
\label{sec:rela}
\subsection{Contrastive Learning (CL)}
Contrastive learning aims to learn representations that can discriminate between different samples, which is achieved by bringing the positive pairs closer together and the negative pairs farther apart. CL first utilizes data augmentation to obtain multi-views, after which Siamese network~\cite{chen2021exploring,grill2020bootstrap} is used to encode the views further.
Prior symbolic works, like MoCo~\cite{he2020momentum,chen2020improved,chen2021empirical}, SimCLR~\cite{chen2020simple}, BYOL~\cite{grill2020bootstrap} and SimSiam~\cite{chen2021exploring}, have made significant model improvements, such as projector, predictor, momentum encoder, memory bank, to address some critical problems in contrastive learning. In the meantime, other efforts have been made to introduce new insights and improvements from an information perspective, which are more relevant to our work. InfoMin~\cite{tian2020makes} has given an insight that good views share the minimum necessary information to perform the downstream tasks well. Therefore, it gives a learned perspective generator and minimizes the mutual information between the two views. Nevertheless, this work only reduces the gap between shared and optimal information while ignoring demands for minimum information.
In addition, H. Wang~\textit{et al.}~\cite{wang2022rethinking} argues that non-shared task-related information can not be ignored and thus retains non-shared usable information by maximizing the mutual information of view and representation.

The relevant works mentioned above only consider the information problem in the model in a one-sided way, without considering it as a whole. In contrast, our work analyzes the holistic aspects of CL and thus proposes new improvements at each stage, with a balance between information compression and expression.

\subsection{Information Bottleneck~(IB) Principle}
Information bottleneck~(IB) principle~\cite{tishby99information,shwartz2017opening,alemi2017vib,voloshynovskiy2020variational} is an approach based on information theory, which formally describes meaningful and relevant information in the data. IB principle states that the model will become more robust to downstream tasks if the obtained representations discard information in the input that is not helpful for the given task. Specifically, given the original data $x$ with the label $y$, the IB principle proposes to learn a compact and informative representation $z$ of data $x$. The objective of the IB principle can be described as follows:
\begin{align}
    \max &\quad I(z;x) \nonumber\\
    s.t. &\quad I_{\phi}(x;z)\le I_c
\end{align}
where $I(\cdot ;\cdot)$ denotes the mutual information between two random variables. $\phi$ denotes the parameters in encoder network. $I_c$ is the upper bound of mutual information $I_{\phi}(x,z)$, indicating the level of information compression. And in the Lagrangian formulation as:
\begin{align}
    \min I_{\phi}(x;z)-\beta I(z;x)
\end{align}
where $\beta$ is the hyper-parameter balancing the compression term $I_{\phi}(x;z)$ and expression term $I(z;x)$.

Following IB, Alemi~\textit{et al.}~\cite{alemi2017vib} presented a variational approximation form for practical computation and optimization. Voloshynovskiy~\textit{et al.}~\cite{voloshynovskiy2020variational} also gave a new perspective that existing unsupervised, supervised methods based on VAE, GAN are products of the IB framework. Zbontar~\textit{et al.}~\cite{zbontar2021barlow} presented the cross-correlation matrix as the similarity metric, resulting in a new CL method, Barlow Twins. They also found that their method is an instantiation of the IB principle. Following Wang~\textit{et al.}~\cite{wang2019deep}, CMIB~\cite{wan2021multi} extended IB to multi-view representation learning by maximizing both the mutual information between learned representation and shared representation as well as view-specific representation while reducing the unnecessary information by minimizing the mutual information between representation and original representation.

Our paper also explores the understanding of CL under the framework of IB principle, pointing out how information compression and expression are implemented and what problems exist in the CL framework. To improve these two information processing stages, we respectively use masking operation and reconstruction task to achieve better performance.

\subsection{Masked Image Modeling~(MIM)}
Self-supervised Learning~(SSL) seeks to alleviate the strong dependence of deep learning on data labels and has done many promising works, but more improvements in performance are needed. The masked autoencoder method has been widely applied to Natural Language Processing (NLP)~\cite{devlin2018bert,brown2020language} and has shown great performance. In the CV field, BEiT~\cite{bao2021beit} quantized image patches as discrete tokens using an off-the-shelf discrete VAE(dVAE) tokenizer~\cite{ramesh2021zero}, then proposes to predict the masked tokens. Recently, masked autoencoder~(MAE)~\cite{he2022masked} and SimMIM~\cite{xie2022simmim} developed the MIM methods~\cite{wei2022masked,bao2021beit,zhou2021ibot}, using vision transformers~(ViT)~\cite{dosovitskiy2020image} backbone to narrow the data distinction between computer vision and natural language. 

These works have found that computer vision data are natural signals with heavy spatial redundancy. A simple masking operation can directly remove the redundant information by discarding pixels with excellent performance, encouraging us to further enhance the CL data augmentation at the pixel level, i.e., masking out patches.

\begin{figure*}[ht!]
    \centering
    \includegraphics[width=1.8\columnwidth]{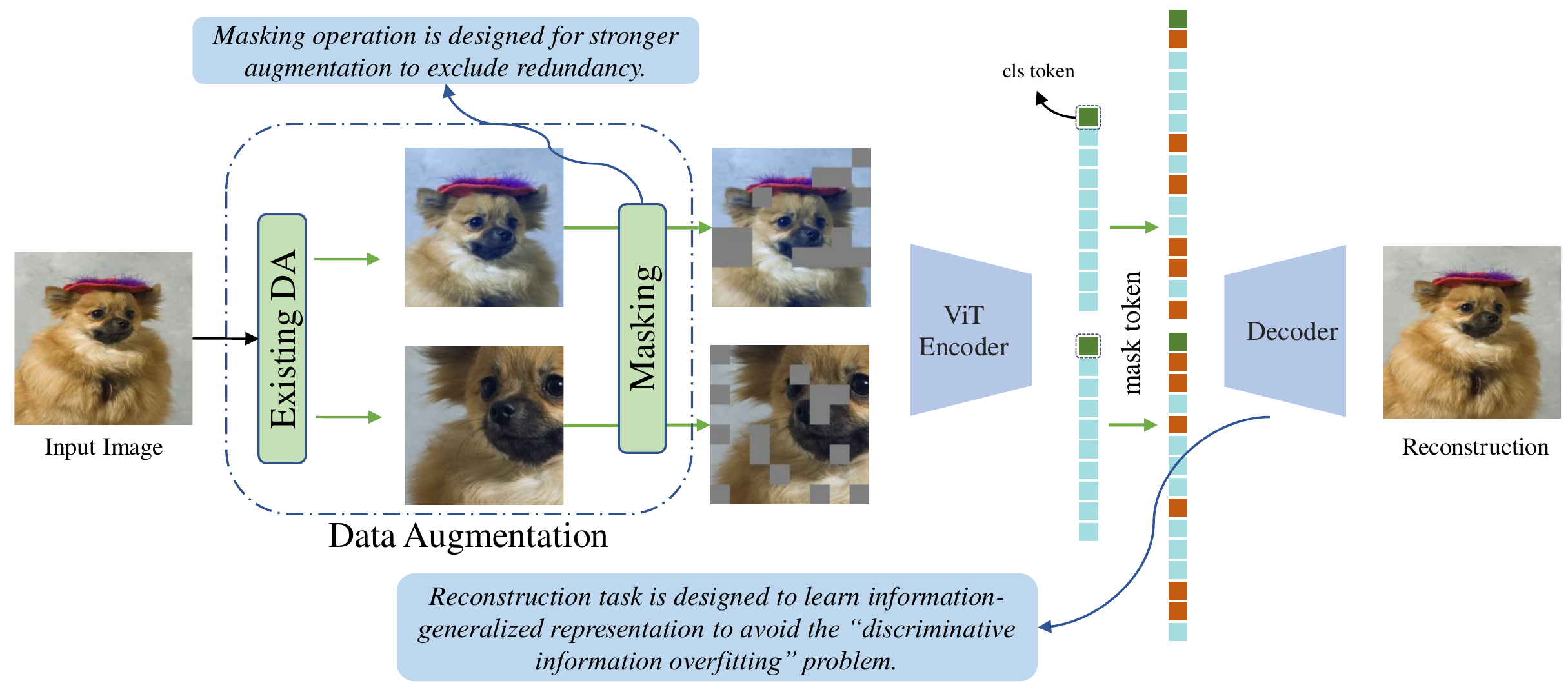}
    \caption{The diagram of our model. We build an encoder-decoder network to learn representation and reconstruct the image. Besides the data augmentation, we further randomly mask out the image patches to reduce the redundant information. Only the unmasked patches are encoded, and then the mask tokens are introduced after the encoder. Both unmasked patches and mask tokens are fed to the decoder to reconstruct the original image.} 
    \label{fig:architecture}
\end{figure*}
\section{Information Analysis and MRCL}
\subsection{Information Analysis on Contrastive Learning}
\label{sec:IBCL}
As illustrated in Section~\ref{sec:intro}, we make a theoretical analysis of CL from an information perspective and find that it is an instantiation of the IB principle. Therefore, this section discusses how information compression and expression work in CL in detail.

For the information compression, both data augmentation (DA)~$\mathcal{T}(\cdot)$ and encoder~$f(\cdot)$ play an essential role, where DA achieves it by different transformations to remove the variance of data, and encoder by siamese network structures and powerful feature extraction capabilities.

Regarding information expression, the discriminative task $y_d$ for positive and negative samples enables the representation to express the discriminative information. Thus we can describe CL as follow:
\begin{align}
    \max &\quad I(z;y_{d}) \label{eq:con_ib} \\ 
    s.t. &\quad I(x;z) = I(x;z_1;z_2)\le I_c \\
    &\quad z_k = f\circ\mathcal{T}_{k}(x), k=1,2
\end{align}
where $I_c$ denotes the compression level. 
In the following, we make a definition of optimal augmentation method and representation with generalized information.

\begin{myDef}
(Sufficient Augmentation) Given an augmentation transformation $\mathcal{T}(\cdot)$, we get an augmentation view $v=\mathcal{T}^{suf}(x)$ of a sample $x$. The augmentation $\mathcal{T}^{suf}(\cdot)$ is called sufficient augmentation for task $y$, if and only if $I(v;y)=I(x;y)$.
\end{myDef}

Intuitively, sufficient augmentation ensures that information about desired downstream tasks $T$ is not removed, thus ensuring that the information in the obtained multi-views works well in the downstream tasks. However, it also remains redundant information that degrades the performance.

\begin{myDef}\label{min_suff_aug}
(Minimum Sufficient Augmentation) The sufficient augmentation is called minimum sufficient augmentation, $\mathcal{T}^{min}$, if and only if its two augmented views satisfy  $I(v_1^{\mathcal{T}^{min}};v_2^{\mathcal{T}^{min}})\le I(v_1^{\mathcal{T}^{suf}};v_2^{\mathcal{T}^{suf}})$, $\forall \mathcal{T}^{suf}$ that is sufficient augmentation.
\end{myDef}

Based on the sufficient augmentation, the minimum sufficient augmentation further removes redundant information (non-task information) to avoid degrading the performance of the downstream tasks. In conclusion, previous works on augmentation are about finding augmentations that satisfy the minimum sufficient augmentation requirement.

\begin{myDef}\label{representation}
(Information-generalized representation) We denote all tasks set as $y$. Thus, the discriminative information over-fitting representation $z_{dis}$ can be expressed as: $I(z_{dis};y)=I(z_{dis};y_{dis})$; The representation is called information-generalized representation $z_{gen}$ if and only if $I(z_{gen};y)\ge I(z_{gen};y_{dis})$, and when the equality holds, it is called the maximal information-generalized representation.
\end{myDef}

Since CL models only seek to discriminate the samples, the \textit{discriminative information overfitting} problem arises, which causes the learned representations to express only discriminative information and thus perform poorly in other downstream tasks. Therefore, to perform well in various downstream tasks, we want the learned information-generalized representation to be enriched with information about more general downstream tasks.
\subsection{A Revisit of Vision Transformers~(ViT)}
Due to the extraordinary performance of vision transformer~(ViT)~\cite{dosovitskiy2020image}, we adopt the ViT backbone in our method. Following the vanilla ViT, we split the image $x\in R^{H\times W\times 3}$ with patch size $P$, yielding a sequence of flattened 2D patches $x_p\in R^{N\times(P^2*3)}$, where $(H;W)$ is the resolution of the original image, $C$ is the number of channels, $N=\frac{HW}{P^2}$ denotes the number of resulting patches. Then we transfer it to $D$ dimensions with the linear projection, as formulated in Eq~\ref{eq:lp}. We refer to the output of this projection as the patch embeddings. Also, we prepend a learnable embedding ($z_0^0=x_{class}$) to the obtained patches embedding. To retain the positional information, positional embeddings~$(\mathbf{E}_{pos})$ are added to the patch embeddings. The resulting sequences are severed to the transformer encoder~\cite{vaswani2017attention}, which consists of $L$ layers of multi-head self-attention~(MSA) and MLP blocks. LayerNorm~(LN) is applied before every block, and residual connections after every block~\cite{baevski2018adaptive,wang2019learning}. The equation is as follow:
\begin{align}
\footnotesize
    z_0&=[x_{class};x^1_{p}\mathbf{E};x^2_{p}\mathbf{E};\cdots;x^N_{p}\mathbf{E}]+\mathbf{E}_{pos}\label{eq:lp}\\
    z^{'}_l&=MAS(LN(z_{l-1}))+z_{l-1},\qquad l=1,2,\cdots,L \label{eq:msa}\\
    z_l& = MLP(LN(z^{'}_l))+z^{'}_l,\qquad l=1,2,\cdots,L \label{eq:mlp}
\end{align}

\subsection{Our Method}
The overall diagram of our model is illustrated in Fig.~\ref{fig:architecture}, which includes three main components. Data augmentation transfers the input image to multi-views, which keeps the image in-variance and removes variance. Unlike previous CL works, we further conduct a masking operation to compress information at the pixel level. The encoder transforms the obtained view into a hidden representation that expresses the extracted data information. Here we adopt ViT, which shows stronger feature extraction capability, as the encoder backbone. The decoder is used to get the image reconstruction, which does not exist in CL models. In this way, we prevent the model from being biased to extract only the discriminative information, thus achieving better downstream tasks performance. Details are given below.
%We now introduce these components in more detail.

Assuming that original data augmentation transfer the input image $x$ into the multi augmented view $\{\hat{v}_k\}_{k=1}^2$, we then get the masked views $\{v_k\}_{k=1}^2$ with a mask ratio $r$. Then we can get the representation $\{z_k\}_{k=1}^2$ of masked view $\{v_k\}_{k=1}^2$.
\paragraph{Masking and Encoder}
Inspired by MIM methods, the masking operation is an efficient way to remove redundant information. As we have analyzed, compressing redundant information is exactly what data augmentation requires. Therefore, after the regular augmentation transformation, we give a masking operation $M_r(\cdot)$ into the masked augmented view~$v$:

\begin{align}
    v_k = M_r(\hat{v}_k) \qquad  i=1,2
\end{align}
where $r$ denotes the mask ratio portion. There exist many masking strategies like 'block-wise'~\cite{bao2021beit}, 'grid-wise' and 'random sampling', where 'random sampling' has shown better performance in many works. Therefore, our paper adopts 'random sampling' as the default masking strategy. The higher the ratio of random sampling is, the more blocks are masked out, i.e., the more redundancy is eliminated. In practice, we first patchify the augmented views $\{\hat{v}\}_{k=1}^2$ and then generate a token for every input patch by linear projection with an added positional embedding. Next, we shuffle the tokens list randomly and drop the last part of the list according to the masking ratio $r$.
For better adaptation to patches masking and better feature information extraction ability, our encoder is ViT~\cite{dosovitskiy2020image} but only applied to the unmasked patches of $\{v_k\}_{k=1}^2$. Following vanilla ViT, our encoder embeds the patches by linear projection, adds positional embedding, and then feeds the resulting patch set into a series of Transformer blocks, obtaining representation $z$.

\begin{table*}[th!]
    \centering
    \small
    \begin{tabular}[width=\textwidth]{l|ccccc}
    \toprule
        Method & Arch &Params.(M) & Pretraining-epochs & Supervision& Top-1 accuracy(\%)   \\
        \midrule
         BYOL\cite{grill2020bootstrap} & ResNet-50 & 23&1000 &RGB&74.3\\
         SimCLR\cite{chen2020simple} & ResNet-50 &23 & 1000&RGB&69.3\\
         BarTwins\cite{zbontar2021barlow} & ResNet-50&23&1000 &RGB&73.2\\
         InfoMin\cite{tian2020makes} & ResNet-50 & 23& 800 &RGB&73.0\\
         MoCo-v2\cite{chen2020improved} & ResNet-50&23&800&RGB&71.1\\
         InfoCL\cite{wang2022rethinking}& ResNet-50&23&200&RGB&61.6\\
         \midrule
         CAE\cite{chen2022context} & ViT-B & 88 & 600 & DALLE+RGB & 68.3\\
         MoCo-v3\cite{chen2021empirical} & ViT-B&88&600&RGB&76.5\\
         DINO\cite{caron2021emerging} & ViT-B & 88 & 300 & RGB & \textbf{\textcolor{blue}{80.1}}  \\
         BEiT\cite{bao2021beit} & ViT-B & 86 & 800  & RGB & 56.7  \\
         SimMIM\cite{xie2022simmim} & ViT-B & 88 & 800 & RGB & 56.7    \\
         iBOT\cite{zhou2021ibot}& ViT-B & 88 & 1600 & RGB & 79.5\\
         MaskFeat\cite{wei2022masked} & ViT-L & 218 & 1600 &HOG& 67.7\\
         MAE\cite{he2022masked} & ViT-L & 218 & 1600 & RGB & 75.1\\
         \midrule[1pt]
         SimCLR(w/ ViT) & ViT-B & 87 & 600 &RGB& 73.1 \\
         \rowcolor{gray!20} MR SimCLR($\alpha=8,\lambda=5$) & ViT-B & 113 & 600 &RGB& $80.0(\uparrow 6.9)$ \\
         \midrule
         BarTwins(w/ ViT) & ViT-B & 87 & 600 &RGB& 74.1\\
         \rowcolor{gray!20} MR BarTwins($\alpha=7,\lambda=2$) & ViT-B & 113 & 600 &RGB& $\textbf{80.4}(\uparrow 6.3)$\\
         \bottomrule
    \end{tabular}
    \caption{Results in ImageNet1K~\cite{deng2009imagenet}. We evaluate our masked reconstruction method with other models, including ResNet and ViT backbones. The \colorbox{gray!20}{Gray} lines denote the results of our MRCL methods, including MR BarTwins and MR SimCLR. The \textbf{Bold} and \textbf{\textcolor{blue}{Blue}} denote the best and second results, respectively.}
    \label{table:Imagenet}
\end{table*}

\paragraph{Reconstruction Decoder}
To alleviate the \textit{discriminative information overfitting} problem, we conduct a decoder to reconstruct views. Although the encoder only encodes the unmasked blocks, the input to the decoder includes encoded visible patches and mask tokens, which are shared, learned vectors indicating the presence of missing patches to be predicted. Moreover, we also add the positional embeddings to the full tokens list to encode the location information. Following MAE, we adopt the ViT as the backbone of the decoder $g(\cdot)$. Due to the independence of the encoder and decoder, it is flexible to design the decoder architecture. In our experiments, we conduct a small and narrow decoder, which forms an asymmetric structure together with the encoder.
Thus, we can get the reconstruction image $g(z)$ with loss:
\begin{align}
    \max\mathcal{L}_{rec}\triangleq\sum_{i=1}^{2}I(z_k;\hat{v}_k)
    \label{con_loss}
\end{align}
Since $I(z_k;\hat{v}_k)=H(\hat{v}_k)-H(\hat{v}_k|z_k)$ and the entropy term $H(\hat{v}_k)$ is is not related to $z_k$, maximizing $I(z_k;\hat{v}_k)$ is equivalent to decreasing the conditional entropy $H(\hat{v}_k|z_k)=-\mathbb{E}_{p(z_k,\hat{v}_k)}[\log p(\hat{v}_k|z_k)]$, i.e, encouraging $z_k$ to express more information about $x$. However the distribution $p(\hat{v}_k|z_k)$ is practically intractable, therefore a hypothetical prior distribution $q(\hat{v}_k|z_k)$ is often used to approximate $p(\hat{v}_k|z_k)$, such as Bernoulli distribution, Gaussian distribution or
Laplace distribution. Here we utilize the Gaussian distribution as the prior, and we therefore get the derivation of Eq.\eqref{con_loss}:
\begin{align}
\small
    \min \sum_{i=1}^2 &H(\hat{v}_k|z_k) \propto\sum_{i=1}^2\mathbb{E}_{p(z_k,\hat{v}_k)}\|\hat{v}_k-g(z_k)\|_2^2-c\triangleq\mathcal{L}_{rec}
\label{eq:rec}
\end{align}
where $Dec(\cdot)$ is the decoder network to compute the mean of Gaussian prior distribution. $c$ is a constant which can be ignored for optimization.

As we described, the patches of reconstruction image $g(z_k)$ include masked and unmasked, which correspond to the prediction and reconstruction task, respectively. Therefore, we seek to explore the respective significance of these two underlying tasks to the model, and Eq.\eqref{eq:rec} can be decomposed as two terms:
\begin{align}
    \mathcal{L}_{rec}=\alpha\mathcal{L}_{rec}^{mask}+\lambda\mathcal{L}_{rec}^{unmask}
\label{eq:rec_de}
\end{align}

\paragraph{Training Objective}
As our analysis in Section.\ref{sec:IBCL}, CL can be written as the form IB framework, Eq.\eqref{eq:con_ib}. Then together with our practical designs of the masking operation $M(\cdot)$ and reconstruction task, we can further derive our MRCL model as:

\begin{align}
    \max &\quad I(z,y_{d})+\sum_{k=1}^2 I(z_k;\hat{v}_k)  \\ 
    s.t. &\quad I(x,z) = I(x;z_1;z_2)\le I_c        \\
    &\quad z_k = f\circ M_r\circ\mathcal{T}_{k}(x), k=1,2
\end{align}

Specifically, after expanding the comparison learning loss term $I(z,y_{d})$ and the reconstruction loss term $\sum_{k=1}^2 I(z_k;\hat{v}_k)$, the model can be derived as:
\begin{align}
     \min \; &\mathbb{E}\left[\log\frac{e^{h(f(\hat{v}_1^i);f(\hat{v}_2^i))}}{\sum_{j=1}^Ne^{h(f(\hat{v}_1^i);f(\hat{v}_2^j))}}\right]\nonumber\\
     &+\sum_{k  =1}^2\mathbb{E}_{p(z_k,\hat{v}_k)}[\alpha\|\hat{v}_k^{mask}-g(z_k)\|_2^2\nonumber\\
     &\qquad\qquad\quad+\lambda\|\hat{v}_k^{unmask}-g(z_k)\|_2^2]
\label{eq:loss}
\end{align}
where $\alpha$ and $\lambda$ are hyper-parameters, weighting the corresponding tasks losses.
The term $\|\hat{v}_k^{mask}-g(z_k)\|_2^2$ denotes the reconstruction loss on masked patches, which is essentially considered a prediction task. The term $\|\hat{v}_k^{unmask}-g(z_k)\|_2^2$ indicates the loss of reconstructing unmasked patches.

\section{Experiments}
In this section, we first verify the effectiveness of our models on ImageNet1K~\cite{deng2009imagenet}. Our method is a refinement of the contrastive learning framework. Therefore, we choose three dominant models: SimCLR~\cite{chen2020simple}, BYOL~\cite{grill2020bootstrap} and Barlow Twins~\cite{zbontar2021barlow} to make improvements for experimental comparison.

\paragraph{Pretraining} Following MoCo~v3~\cite{chen2021empirical}, we adopt the AdamW optimizer as default, and the momentum is set to $\beta=(0.9,0.95)$. Besides, the weight decay is set to $1\mathrm{e}^{-6}$ and the linear scaling rule~\cite{goyal2017accurate}: $lr = base\_lr\times batch\_size/256$ is introduced to set the learning rate. The base learning rate $base\_lr$ is $1.5\mathrm{e}^{-4}$ with the batch size of $512$. Note that cosine learning rate schedule~\cite{loshchilov2016sgdr} with a warmup~\cite{goyal2017accurate} is adopted for early 30 epochs. All pre-training experiments are conducted on 8 NVIDIA RTX3090 GPUs.

\begin{table*}[t!]
    \centering
    \small
    \renewcommand\arraystretch{1.1}
    \begin{subtable}[h]{0.42\textwidth}
        \setlength{\tabcolsep}{6pt}
        \centering
        \begin{tabular}{l|cccc}
    \toprule
        Method & Pre-Epoch & mIoU  \\
        \midrule
        MoCo-v3\cite{chen2021empirical} & 300 & 47.3\\
        BEiT\cite{bao2021beit} & 800 & 47.1\\
        iBOT\cite{zhou2021ibot} & 1600 & \textbf{\textcolor{blue}{50.0}}\\
        CAE\cite{chen2022context} & 1600 & 44.0\\
        DINO\cite{caron2021emerging} & 400 & 47.2\\
        MAE\cite{he2022masked} & 1600 & 48.1\\
        \midrule
        MR BarTwins & 600 & \textbf{51.5}($\uparrow 1.5$)\\
        MR SimCLR & 600 & \textbf{51.3}($\uparrow 1.3$)\\
        \bottomrule
    \end{tabular}
      \caption{Semantic segmentation results on ADE20K. Upernet~\cite{xiao2018unified} is used as the default segmentation framework.}
      \label{tab:ADE20K}
    \end{subtable}
    \hfill
    \begin{subtable}[h]{0.54\textwidth}
        \centering
        \setlength{\tabcolsep}{8pt}
        \begin{tabular}{l|ccccc}
    \toprule
         Method & Epochs & $AP^{bbox}$ & $AP^{mask} $  \\
         \midrule
         MoCo-v3\cite{chen2021empirical} & 300 & 47.9 & 42.7\\
         BEiT\cite{bao2021beit} & 800 & 49.8 & 44.4\\
         iBOT\cite{zhou2021ibot} & 1600 & 51.2 & 44.2 \\
         CAE\cite{chen2022context} & 1600 & 50.1 & 44.0\\
         SIM\cite{tao2022siamese} & 1600 & 49.1 & 43,8\\
         MAE\cite{he2022masked} & 1600 & \textbf{\textcolor{blue}{52.4}} & \textbf{\textcolor{blue}{46.5}}\\
         \midrule
         MR BarTwins & 600 & \textbf{53.3}$(\uparrow 0.9)$& \textbf{46.6}$(\uparrow 0.1)$\\
         MR SimCLR & 600 & \textbf{53.7}$(\uparrow 1.3)$ & \textbf{46.9} $(\uparrow 0.4)$\\
         \bottomrule
    \end{tabular}
        \caption{COCO object detection and segmentation. We use the Mask R-CNN model~\cite{he2017mask} as our framework.}
        \label{tab:COCO}
     \end{subtable}
     \caption{Results on the transfer experiments including semantic segmentation
    and object detection. MR denotes the model that merges our method. The \textbf{Bold} indicates the best result and \textbf{\textcolor{blue}{Blue}} denotes the second place.}
     \label{tab:transfer_exp}
\end{table*}

\subsection{Results on ImageNet1K}
Following previous works~\cite{caron2021emerging,chen2021empirical,chen2020simple,bao2021beit}, we evaluate MRCL in ImageNet1K~\cite{li2021image} linear-probing. After pre-training, we follow~\cite{chen2021empirical} to use SGD optimizer for training 100 epochs. The learning rate is 0.2 with the batch size of $1024$ and $wd$ of 0. We only use random resized cropping and flipping augmentation. The classification top-1 accuracy results are shown in Table.~\ref{table:Imagenet}. Many typical and widely applied works, including works with the ResNet and ViT backbones, are in the comparison. We utilize two classical models as our baseline, BarTwins, and SimCLR (both with ViT-B backbone), and they are denoted as MR BarTwins and MR SimCLR.

As shown in Table.\ref{table:Imagenet}, MR BarTwins model achieves the state-of-the-art result $80.4\%$, which is $0.3\%$ higher than DINO~\cite{caron2021emerging}, indicating that improvements in information compression and expression can extract better information content to serve downstream classification tasks.

When $\alpha=8,\lambda=5$, our MR SimCLR achieves $80.0\%$ top-1 linear probing accuracy(\colorbox{gray!20}{Gray} in Table.\ref{table:Imagenet}), with $6.9\%$ better than vanilla SimCLR (with ViT-B backbone) . When $\alpha=7,\lambda=2$, MR BarTwins achieve the $80.4\%$ accuracy (\colorbox{gray!20}{Gray} in Table.\ref{table:Imagenet}), which makes $6.3\%$ improvement for original BarTwins model (with ViT-B backbone). These remarkable improvements are due to adding our masking and reconstruction tasks to the underlying models, allowing it to learn more valuable information. In addition, $\alpha\ge\lambda$ means that the pixel prediction task contributes more than reconstruction, i.e., prediction can provide more information, which is also demonstrated in MAE~\cite{he2022masked} and SimMIM~\cite{xie2022simmim}.

%这里应该着重在强调一下，为什么我们的方法可以带来显著的性能提升，最好是结合不同损失函数的性质来讲，如果没法靠在一起，也要讲的尽可能详细一点，目前下面这段的解释太薄了。本身就是偏理论的文章，所以不能让审稿人觉得我们的结果就是靠实验来说明的。
% All these results illustrate the enhancement effect on classification performance of our method by removing redundant information and extracting essential information.

\subsection{Transfer Results}
To further validate the transferability of our model, we follow previous methods to evaluate pretrained models on ADE20K~\cite{zhou2019semantic} semantic segmentation and MSCOCO~\cite{lin2014microsoft} object detection and segmentation. All models in the comparison adopt ViT-B backbone.

\paragraph{Semantic segmentation} We evaluate our MR BarTwins and MR SimCLR models on ADE20K dataset~\cite{zhou2019semantic}, which includes 25,562 images with 150 semantic categories. By default, Upernet~\cite{xiao2018unified} is used as the segmentation framework. Following the standard setting, the Mean Intersection over Union (mIoU) results are shown in Table.\ref{tab:ADE20K}. MR BarTwins and MR SimCLR achieve $51.5$ and $51.3$ mIoU, respectively, both surpassing the previous best result iBOT~\cite{zhou2021ibot}~($50.0$). This excellent performance on ADE20K demonstrates that our method effectively alleviates the \textit{discriminative information overfitting} problem, allowing the model to learn more generalized information that can be well applied to the semantic segmentation task with good performance.
\paragraph{Objective detection and segmentation} Our MRCL methods conduct the objective detection and segmentation experiments on MSCOCO dataset~\cite{lin2014microsoft}, which consists of $118k$ training, $5k$ validation, and $20k$ test-dev images. MRCL adopts the Mask-RCNN~\cite{he2017mask} as our default objection detection framework. As shown in Table.\ref{tab:COCO}, MRCL methods significantly improve the performance on MSCOCO object detection and segmentation. In the objective detection, MR SimCLR achieves the best results with $1.3$ improvement on $AP^{bbox}$ than MAE~\cite{he2022masked} ($53.7~vs.~52.4$). MR BarTwins also get competitive results of $52.3$. In the objective segmentation, MR SimCLR significantly improves $AP^{mask}$ over MAE by 0.4 points ($46.9~vs.~46.5$). These promising results again verify the effectiveness of our method in various downstream tasks, which benefited from our regularization reconstruction task.
\subsection{Analytical Experiments}
Here, we provide some analytical experiments to understand our model design further. The analytical experiments are performed on STL-10 dataset~\cite{coates2011analysis} and the base models are BarTwins~\cite{zbontar2021barlow} and BYOL~\cite{grill2020bootstrap}. We random crop and resize the STL-10 data to $224\times 224$. We also adopt the AdamW optimizer, and the base learning rate is $3\mathrm{e}^{-4}$ with a batch size of $512$. The default training epochs are set to 300. Following experiments on ImageNet, we adopt a learning rate warmup for 30 epochs. After the warmup, the learning rate follows a cosine decay schedule~\cite{loshchilov2016sgdr}.

\begin{figure}[h!]
    \centering
    \includegraphics[width=0.5\textwidth]{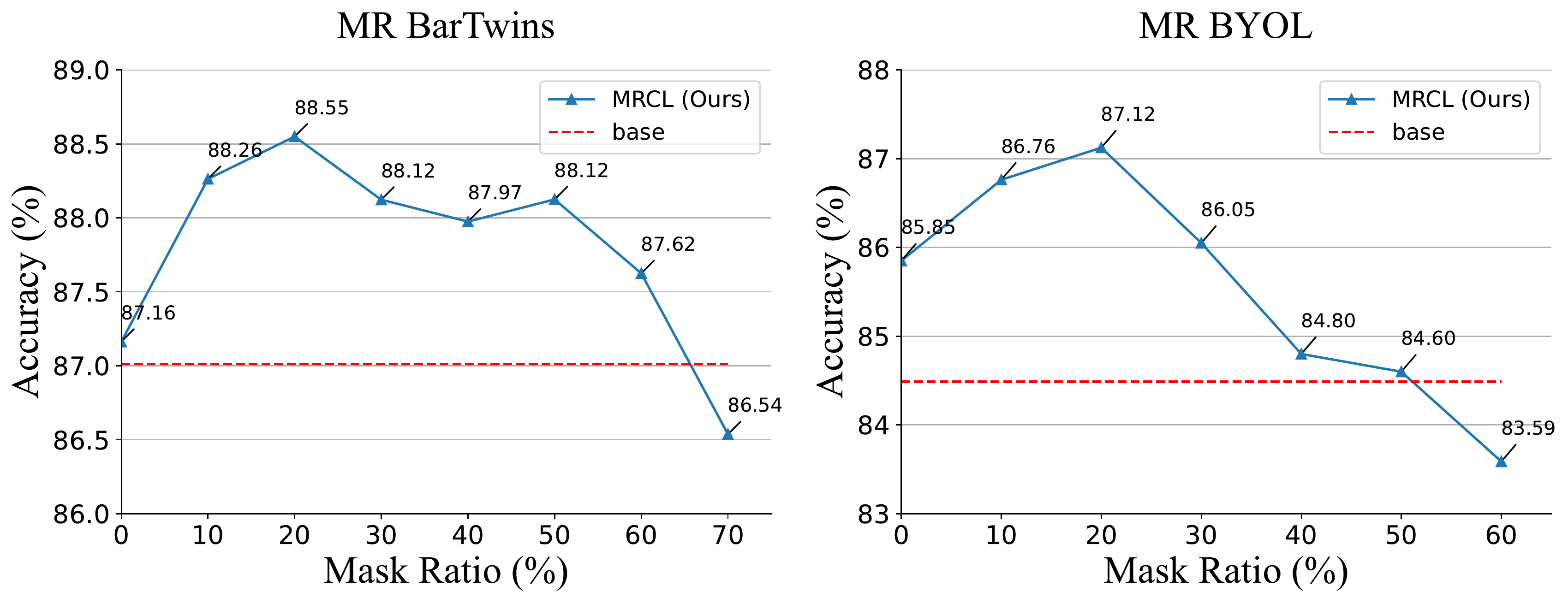}
    \caption{The effect of mask ratio. The red dotted line denotes the accuracy of baseline model.}
    \label{fig:ratio_abl}
\end{figure}
\paragraph{Masking Ratio} According to our analysis based on the information bottleneck principle, mask ratio indicates the degree of information compression: the larger the ratio, the more compression is obtained, and vice versa the smaller. We fix the reconstruction hyperparameter $\alpha=\beta=10$ and $\alpha=\beta=0.01$ in MR BarTwins and MR BYOL, respectively. The results are present in Fig.~\ref{fig:ratio_abl}, where $r=0$ denotes that we only add the reconstruction task to the base model without masking. When $r=0$, both models show an accuracy improvement, indicating that reconstruction regularization provides an excellent alleviation for the \textit{discriminative information overfitting} problem. The MR BarTwins and MR BYOL models show the best result when $r=20\%$ and the whole accuracy curve shown in Fig.~\ref{fig:ratio_abl} shows a reversed U-shape. This indicates that there is still redundancy information after the basic augmentation ($r=0$), thus reducing the performance. Moreover, overly strong compression (higher mask ratio) leads to loss of valuable information and thus degrades performance too. 

\begin{figure}[h!]
    \centering
    \includegraphics[width=0.5\textwidth]{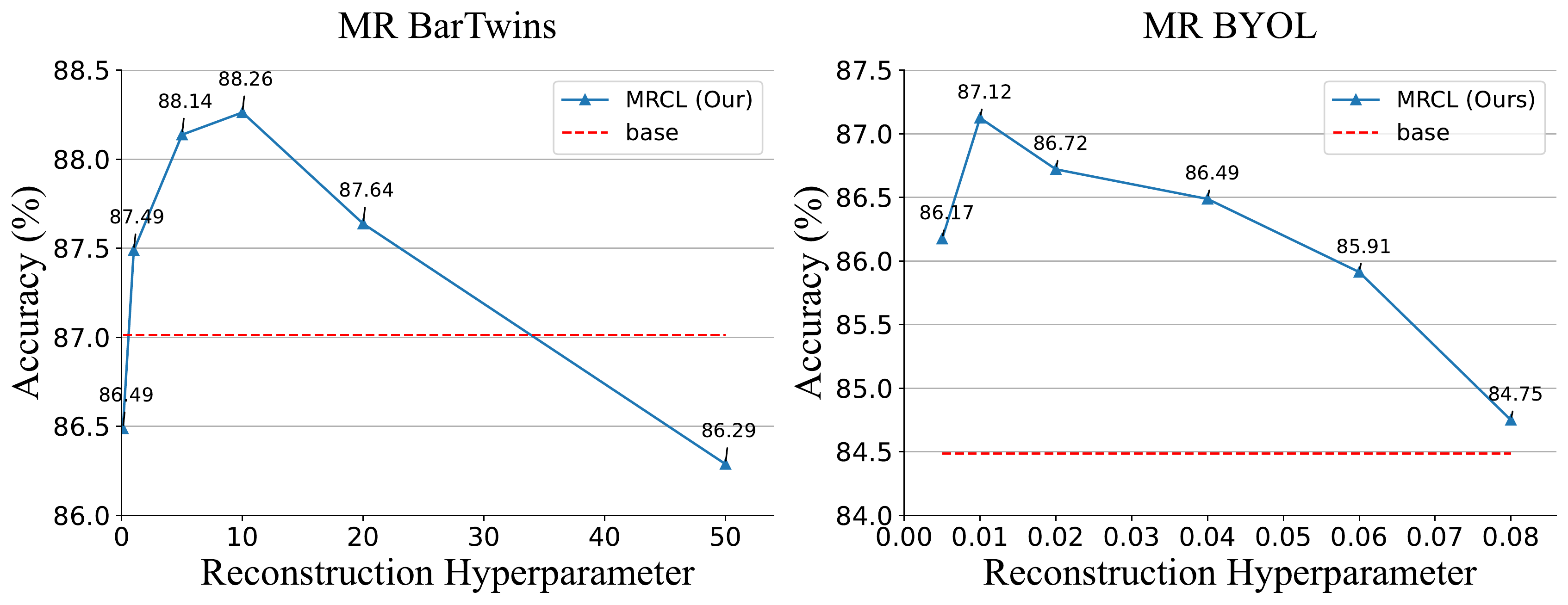}
    \caption{Ablation on the hyperparameter of reconstruction loss. The red dotted line denotes the accuracy of baseline model.}
    \label{fig:abl2}
\end{figure}

\paragraph{Reconstruction Weight}
Our MRCL method introduces a reconstruction task to lead the model to express more essential information. Here we ablate the weight of the reconstruction loss to explore the importance of the task. The mask ratio is fixed at $0.1$ and $0.2$, respectively, in MR BarTwins and MR BYOL. The results in Fig.~\ref{fig:abl2} show that when $\alpha=\beta=10$, MR BarTwins get the best performance, which was $1.25\%$ higher than the baseline model~(w/o reconstruction task). MR BYOL achieves the best performance when $\alpha=\beta=0.01$, with $2.6\%$ higher than the base model. Both models have minimum and maximum thresholds beyond which performance degradation can occur. A small reconstruction hyperparameter force the model to learn based on the discriminative task, which still suffers from the discriminative information overfitting problem. In the meantime, a larger reconstruction parameter makes the model entirely biased to reconstruct pixels, ignoring the extraction of semantic information from the image and degrading the performance.

\begin{figure}[h!]
    \centering
    \includegraphics[width=0.5\textwidth]{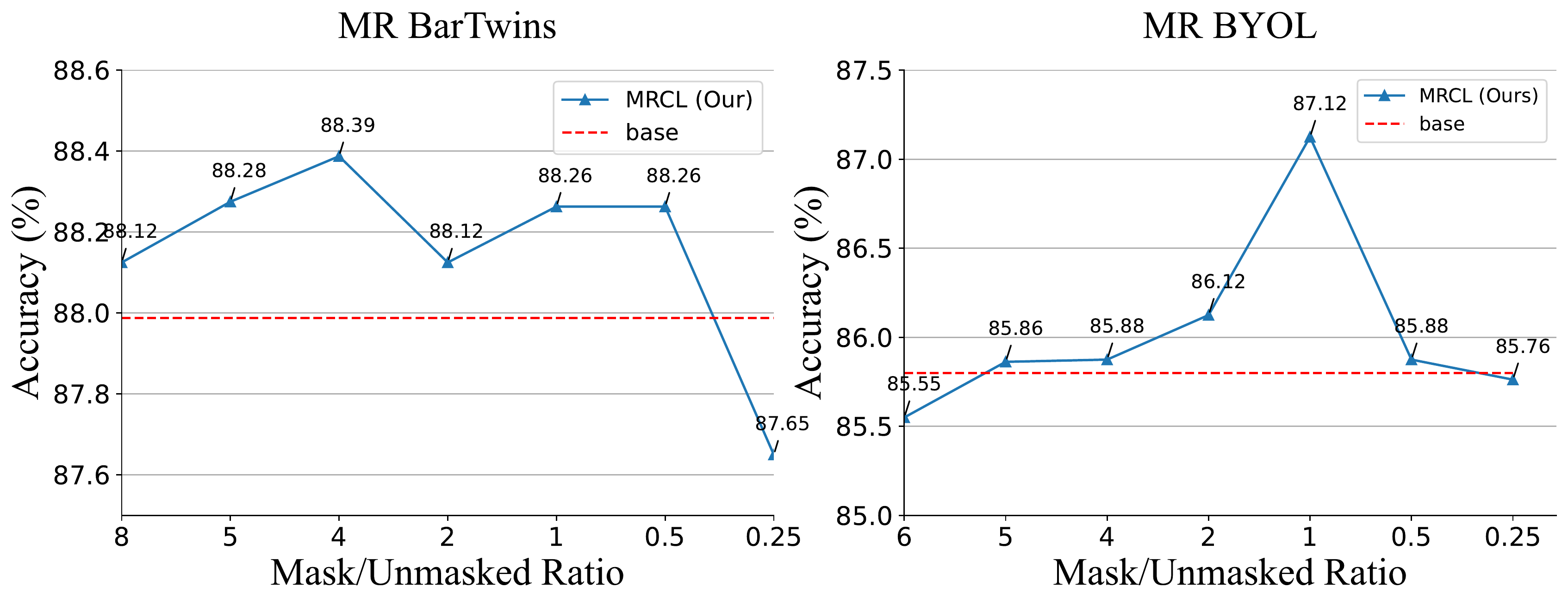}
    \caption{Exploratory experiments on the importance of masked and unmasked patches. The red dotted line denotes the accuracy of baseline model.}
    \label{fig:abl3}
\end{figure}
\paragraph{Reconstruction vs. Prediction}
As we described, the reconstruction loss can be decomposed as Eq.~\eqref{eq:rec_de}, including unmasked patches reconstruction and masked patches prediction. Here we ablate the weight of these two different tasks to improve the model, with fix the mask ratio $r=0.1~(0.2)$ and masked patches weight $\alpha=10~(0.01)$ in MR BarTwins (MR SimCLR). We get different $mask/unmask$ ratios by adjusting the unmasked patches weight $\beta$. $\alpha/\beta\ge 1$ indicates more focus on prediction tasks, while $\alpha/\beta < 1$ indicates more attention on reconstruction.
Results in Fig.~\ref{fig:abl3} show that more mask weighting (i.e., $mask/unmask=4$ in MR BarTwins and $mask/unmask=1$ in MR BYOL, respectively) gives the best results. However, an upper limit exists beyond which the model learning becomes completely biased towards prediction, leading to greater difficulty and performance degradation. Also, this mask ratio has a lower bound, namely the relatively excessive weight of unmasked patches. In this case, the model collapses to a pure reconstruction task, which loses information from the unmasked patches, leading to poor results. These result curves also show a reversed U-shape, demonstrating our minimum sufficient information statement.
\section{Discussion and Conclusion}
As a critical problem of self-supervised learning, contrastive learning has become significant in deep learning due to its non-dependence on labels and good performance in downstream tasks. However, its discriminative task setting leads to the \textit{discriminative information overfitting} problem in the learned representation, degrading performance in various downstream tasks. Here we first make a theoretic analysis of the contrastive learning framework via the IB principle, pointing out the information compression and expression process. We then make the following improvements in the respective process: for better information compression, we utilize the masking operation to achieve stronger data augmentation, thus removing unnecessary and redundant information; for resolving the \textit{discriminative information over-fitting} problem to achieve better expression, we propose to add a regularization reconstruction task to learn information-generalized representation. In practice, the state-of-the-art results in ImageNet1K image classification, ADE20K semantic segmentation, and COCO objective detection and segmentation illustrate that the MR approach practically implements our analysis based on the IB principle.

%-------------------------------------------------------------------------
%-------------------------------------------------------------------------
%-------------------------------------------------------------------------
%------------------------------------------------------------------------
% \newpage
%%%%%%%%% REFERENCES
%{\small
%\bibliographystyle{ieee_fullname}
%\bibliography{egbib.bib}
%}

\end{document}